\documentclass[11pt]{article}

\usepackage[final]{acl}

\usepackage{times}
\usepackage{latexsym}

\usepackage[T1]{fontenc}

\usepackage[utf8]{inputenc}

\usepackage{microtype}

\usepackage{inconsolata}

\usepackage{graphicx}

\usepackage{times}
\usepackage{latexsym}
\usepackage[T1]{fontenc}
\usepackage[utf8]{inputenc}
\usepackage{microtype}
\usepackage{graphicx}
\usepackage{amsmath}
\usepackage{amssymb}
\usepackage{algorithm}
\usepackage{algorithmic}
\usepackage{booktabs}
\usepackage{multirow}
\usepackage{xcolor}

\title{From Logs to Language: Learning Optimal Verbalization for LLM-Based Recommendation at Industry Scale }

\author{
  {\bf Yucheng Shi}$^{1,}$\thanks{Work done at Netflix.}, 
  {\bf Ying Li}$^2$, 
  {\bf Yu Wang}$^{3,\ast}$, 
  {\bf Yesu Feng}$^2$, 
  {\bf Arjun Rao}$^2$, 
  {\bf Rein Houthooft}$^2$, \\
  {\bf Shradha Sehgal}$^2$, 
  {\bf Jin Wang}$^2$, 
  {\bf Hao Zhen}$^1$, 
  {\bf Ninghao Liu}$^4$, 
  {\bf Linas Baltrunas}$^2$ \\
  \textnormal{$^1$University of Georgia \quad $^2$Netflix \quad $^3$Capital One \quad $^4$Hong Kong Polytechnic University}
}

\begin{document}
\maketitle

\begin{abstract}
Large language models (LLMs) are promising backbones for generative recommender systems, yet a key challenge remains underexplored: \emph{verbalization}, i.e., converting structured user interaction logs into effective natural language inputs. Existing methods rely on rigid templates that simply concatenate fields, yielding suboptimal representations for recommendation. We propose a data-centric framework that \emph{learns} verbalization for LLM-based recommendation. Using reinforcement learning, a verbalization agent transforms raw interaction histories into optimized textual contexts, with recommendation accuracy as the training signal. This agent learns to filter noise, incorporate relevant metadata, and reorganize information to improve downstream predictions. Experiments on a large-scale industrial streaming dataset from \emph{Netflix} show that learned verbalization delivers up to \textbf{93\%} relative improvement in discovery item recommendation accuracy over template-based baselines. Further analysis reveals emergent strategies such as user interest summarization, noise removal, and syntax normalization, offering insights into effective context construction for LLM-based recommender systems.

\end{abstract}

\section{Introduction}

The integration of large language models into recommendation systems has garnered significant attention, with approaches ranging from using LLMs as feature extractors to deploying them as end-to-end reasoning engines \citep{wu2024survey, fan2023recommender, lin2023can, firooz2025360brew}. A critical yet underexplored component in this paradigm is \emph{verbalization}, the process of converting structured user behavioral data into natural language that LLMs can process. Current approaches predominantly rely on template-based methods that mechanically concatenate interaction fields, assuming that LLMs can effectively parse and reason over such representations.

We argue that this assumption fundamentally misaligns with how LLMs process information. Consider a typical user interaction log containing timestamps, device, item IDs, engagement types, and viewing durations. While such granular data is valuable for traditional collaborative filtering approaches, presenting it verbatim to an LLM introduces several challenges. First, the heterogeneous feature space creates parsing overhead that consumes model capacity. Second, not all interactions carry equally important predictive signal. Finally, the absence of semantic context (e.g., metadata information) forces the LLM to reason purely from behavioral patterns without access to content understanding, which harms the recommendation quality especially for cold-start items.

These observations motivate a fundamental question: \emph{Can we learn to verbalize user interaction data specifically for LLM-based recommender, rather than treating verbalization as a fixed preprocessing step?}

This paper introduces a \textbf{data-centric} framework for LLM-based recommendation that treats verbalization as a learnable component. Our approach decomposes the recommendation pipeline into two specialized components. The \textbf{Verbalizer} is a generative model that transforms raw user interaction sequences into optimized natural language descriptions, learning to filter noise, add relevant metadata, and restructure information for downstream reasoning. The \textbf{Reasoner} is a causal language model that performs the actual recommendation task, predicting user preferences from the verbalized context.

The key insight enabling this decomposition is that verbalization quality can be measured through recommendation accuracy. If a verbalized representation leads to correct predictions, it has successfully preserved and highlighted the most relevant preference signals. This creates a natural training signal: we can optimize the Verbalizer using the Reasoner's prediction accuracy as reward. We instantiate this idea through Group Relative Policy Optimization (GRPO), adapting the algorithm for both Verbalizer training (where the Reasoner provides rewards) and subsequent Reasoner training (where ground-truth engagement labels provide rewards).

Our contributions are fourfold. First, we formalize the verbalization optimization problem for LLM-based recommendation and propose a two-component architecture that separates context optimization from preference reasoning. Second, we develop a two-stage GRPO-based training framework that enables end-to-end optimization of both components without requiring explicit verbalization labels. Third, through extensive experiments on industrial-scale data, we demonstrate up to 93\% relative improvement in recommendation accuracy. Lastly, we discovered several insightful patterns in verbalization optimization for LLM-based recommenders.

\section{Related Work}

\paragraph{LLMs for Recommendation.} Recent work has explored multiple paradigms for incorporating LLMs into recommendation systems. One line of research uses LLMs to generate textual features or embeddings that augment traditional recommendation models \citep{li2023text, hou2024large, bao2023tallrec}. Another approach employs LLMs directly as recommenders through in-context learning, where user history and candidate items are formatted as prompts \citep{wang2023llm4rec, dai2023uncovering, liu2023chatgpt}. \citet{geng2022recommendation} proposed P5, a unified text-to-text framework that formulates various recommendation tasks as natural language generation. \citet{zhang2023recommendation} provided a comprehensive survey on the intersection of LLMs and recommender systems. More recently, \citet{zheng2024adapting} explored parameter-efficient fine-tuning methods for adapting LLMs to recommendation tasks, while \citet{liao2024llara} proposed hybrid approaches combining LLM reasoning with traditional collaborative signals. Our work falls into the generative recommendation category but introduces a crucial distinction: rather than treating prompt construction as a fixed engineering choice, we learn the verbalization function end-to-end.

\paragraph{Prompt Optimization.} Prompt engineering and optimization have been widely studied in NLP \citep{liu2023pre, zhou2023large}, spanning discrete prompt search \citep{shin2020autoprompt, gao2021making}, continuous prompt tuning \citep{lester2021power, li2021prefix}, and reinforcement learning–based methods \citep{deng2022rlprompt, zhang2023tempera}. Recent work further explores gradient-based optimization \citep{pryzant2023automatic} and LLM-driven prompt optimization \citep{yang2024large}. These advances motivate our work: if prompts can be optimized for task performance, then the verbalization of user interaction data, the core input to LLM-based recommenders, should also be optimizable. However, existing methods focus on task-level optimization with fixed input-output mappings. In contrast, recommendation requires user-dependent, instance-level verbalization. Our work extends prompt optimization to learn instance-specific verbalizations that directly maximize recommendation accuracy.

\paragraph{Reinforcement Learning for LLMs.} Our training methodology builds upon recent advances in RL-based LLM alignment. Proximal Policy Optimization (PPO) has been widely adopted for RLHF \citep{ouyang2022training}, though its requirement for a learned value function introduces computational overhead. Group Relative Policy Optimization (GRPO) offers an efficient alternative by estimating advantages from group-level reward statistics \citep{shao2024deepseekmath}, which we adapt for both verbalization and reasoning optimization.

\section{Problem Formulation}

We consider the sequential recommendation setting where each user $u$ is associated with an interaction history $\mathcal{H}_u = \{h_1, h_2, \ldots, h_T\}$. Each interaction $h_t$ is a structured record containing heterogeneous features: timestamp $\tau_t$, content identifier $c_t$, engagement type $e_t$ (e.g., play, thumb up, add to list), viewing duration $\delta_t$. The recommendation task is to predict the next item $y^*$ the user will engage with from a candidate set $\mathcal{C} = \{y_1, y_2, \ldots, y_N\}$.

A basic template-based verbalization constructs a textual representation through a deterministic mapping $\phi_{\text{template}}: \mathcal{H}_u \rightarrow \mathcal{X}$, where $\mathcal{X}$ denotes the space of natural language strings. For example:

\begin{quote}
\small
\texttt{20250608, Monday, 2 PM, ID: 123456, Stranger Things, Play, Duration: 80.08 min}
\end{quote}

We propose learning a parameterized verbalization function $\phi_\theta: \mathcal{H}_u \rightarrow \mathcal{X}$ that produces optimized verbalizations. The optimization objective is to maximize recommendation accuracy:
\begin{equation}
\max_\theta \mathbb{E}_{u, \mathcal{C}, y^*}\left[\mathbf{1}\left[\text{Reasoner}(\phi_\theta(\mathcal{H}_u), \mathcal{C}) = y^*\right]\right]
\end{equation}
where $\text{Reasoner}(\cdot, \cdot)$ denotes the prediction from the reasoning model given verbalized context and candidates.

\section{Method}

\begin{figure*}[t]
\centering
 \includegraphics[width=0.9\linewidth]{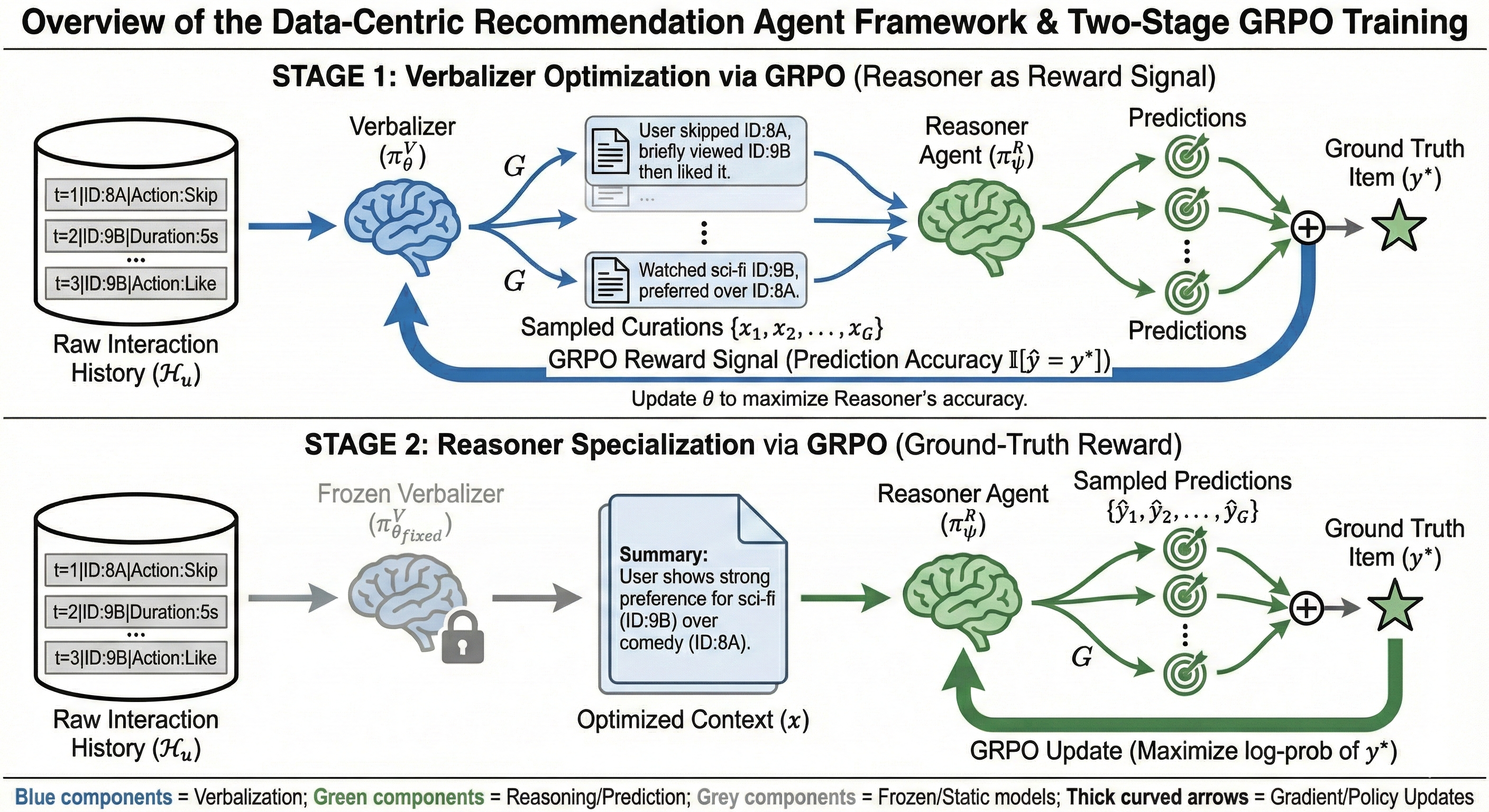}
\caption{Two-stage training pipeline for the Data-Centric Recommendation Agent.}
\label{fig:architecture}
\end{figure*}

Our framework consists of two components, the Verbalizer and the Reasoner, trained through a two-stage pipeline using variants of Group Relative Policy Optimization. We first describe the component architectures, then detail the training procedures.

\subsection{Verbalizer}

The Verbalizer $\pi_\theta^V$ is a generative language model that takes raw interaction history as input and produces an optimized textual representation. We investigated two architectural variants for the verbalization process.

\paragraph{Action-Based.} This variant outputs discrete decisions for each interaction: 1) Filter: whether to keep or discard an interaction; 2) Add: whether to enrich it with metadata. Formally, for each $h_t \in \mathcal{H}_u$, the model generates a binary keep decision $k_t \in \{0, 1\}$ and an enrichment decision $m_t \in \{0, 1\}$. The output takes the form:
\begin{equation}
\texttt{time, ID, } k_t \texttt{, } m_t \texttt{,}
\end{equation}
These decisions are then executed deterministically: kept interactions with $m_t = 1$ are augmented with additional metadata.

\paragraph{Rewrite-Based.} This variant generates a complete textual rewrite of the interaction history, providing maximum flexibility for the model to aggregate, summarize, and restructure information. The model is prompted to transform interactions into natural language while preserving important signals and removing noise.

Our experiments reveal that the rewrite-based approach substantially outperforms action-based verbalization, as it enables emergent behaviors such as aggregating repetitive patterns (``watched 5 episodes of X'') and generating member interest summaries (``showed strong preference for dark thrillers''). We adopt the rewrite-based architecture for our primary experiments.

\subsection{Reasoner}

The Reasoner $\pi_\psi^R$ is a causal language model that receives the verbalized context $x = \phi_\theta(\mathcal{H}_u)$ and candidate set $\mathcal{C}$, and outputs a prediction over candidates. The model generates its prediction through standard autoregressive decoding, with the predicted item extracted from the generated text.

\subsection{Stage 1: Verbalizer Optimization via GRPO}
\label{sec:verbalizer_grpo}

Training the Verbalizer presents a fundamental challenge: no ground-truth ``optimal verbalization labels'' exist. We address this by treating this as a reinforcement learning problem where a Reasoner provides reward signals. Intuitively, if a verbalization leads to correct recommendation, it has successfully captured the relevant preference information.

\paragraph{Oracle Reasoner for Reward Signal.} A critical design choice is which model provides the reward signal during Verbalizer training. We employ a capable closed-source LLM (rather than the target Reasoner being trained) as the oracle Reasoner for several reasons. First, more powerful models provide higher-quality reward signals with better discrimination between effective and ineffective verbalizations. Second, using a fixed oracle avoids the instability of co-training both components simultaneously. Third, the oracle's superior reasoning capabilities help the Verbalizer learn verbalizations that capture genuine preference signals rather than exploiting weaknesses in a weaker Reasoner. Empirically, we find that training with a stronger oracle leads to verbalizations that generalize better to different Reasoner architectures.

We employ Group Relative Policy Optimization (GRPO) for Verbalizer training. For each user $u$ with history $\mathcal{H}_u$, candidates $\mathcal{C}$, and ground truth $y^*$, we sample a group of $G$ verbalizations from the current policy:
\begin{equation}
\{x_1, x_2, \ldots, x_G\} \sim \pi_\theta^V(\cdot | \mathcal{H}_u).
\end{equation}
Each verbalization $x_i$ is evaluated by passing it through the oracle Reasoner to obtain a prediction, yielding reward:
\begin{equation}
r_i = \begin{cases}
1 & \text{if } \pi_{\text{oracle}}^R(x_i, \mathcal{C}) = y^* \\
0 & \text{otherwise}
\end{cases}
\end{equation}

The rewards are normalized within each group to compute advantages:
$\{
\hat{A}_i = \frac{r_i - \text{mean}(\mathbf{r})}{\text{std}(\mathbf{r}) + \epsilon}
\}$
where $\mathbf{r} = \{r_1, \ldots, r_G\}$ and $\epsilon$ is a small constant for numerical stability.
The Verbalizer policy is updated by maximizing the GRPO objective:
\begin{equation}
\begin{split}
\mathcal{J}^V(\theta) = \mathbb{E}\Bigg[ \frac{1}{G}\sum_{i=1}^G \frac{1}{|x_i|} \sum_{t=1}^{|x_i|} \Big\{ \min \Big[ \rho_{i,t} \hat{A}_i,  \\
\text{clip}(\rho_{i,t}, 1-\epsilon_v, 1+\epsilon_v) \hat{A}_i \Big] - \beta_v \mathcal{D}_{\text{KL}} \Big\} \Bigg]
\end{split}
\label{eq:verbalizer_grpo}
\end{equation}
where $\rho_{i,t} = \frac{\pi_\theta^V(x_{i,t} | \mathcal{H}_u, x_{i,<t})}{\pi_{\theta_{\text{old}}}^V(x_{i,t} | \mathcal{H}_u, x_{i,<t})}$ is the importance sampling ratio, $\epsilon_v$ is the clipping parameter, and $\beta_v$ weights the KL divergence regularization from a reference policy.

\paragraph{Length Reward.} Unconstrained optimization may lead to degraded solutions, either extremely compressed outputs that lose information or verbose outputs that introduce high training and inference costs. We incorporate a length-based reward component:
\begin{equation}
r_i^{\text{total}} = \alpha \cdot r_i^{\text{acc}} + (1-\alpha) \cdot r_i^{\text{len}}
\end{equation}
where $r_i^{\text{len}}$ is a plateau function that provides maximum reward for compression ratios in a target range (e.g., 0.3–0.7), with penalties for both over-compression and under-compression. We set $\alpha = 0.9$ in our experiments.

\begin{figure}[t]
\centering
\includegraphics[width=0.95\linewidth]{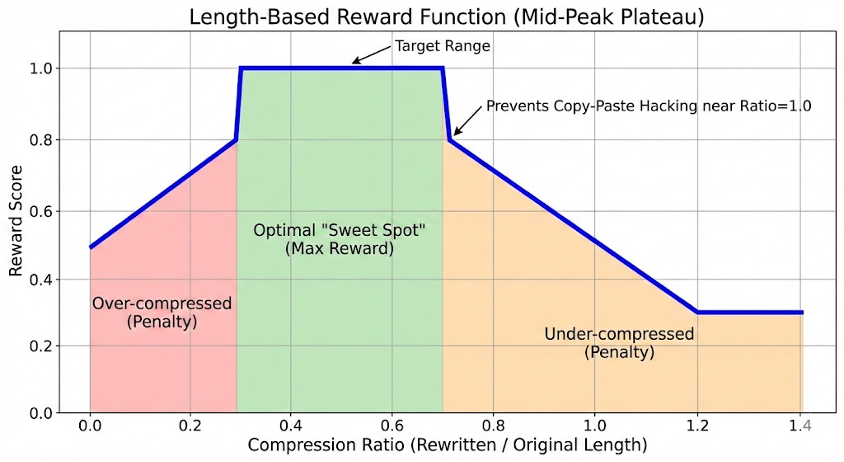}
\caption{Length reward function design.}
\label{fig:length_reward}
\vspace{-10pt}
\end{figure}

\subsection{Stage 2: Reasoner Optimization via GRPO}
\label{sec:reasoner_grpo}

After Verbalizer training converges, we freeze the Verbalizer and train the Reasoner on the verbalized data distribution. This stage adapts the Reasoner to the specific verbalization patterns produced by the trained Verbalizer.
For each training instance, the frozen Verbalizer generates a single verbalization $x = \phi_\theta(\mathcal{H}_u)$. The Reasoner then samples $G$ predictions:
$
\{\hat{y}_1, \hat{y}_2, \ldots, \hat{y}_G\} \sim \pi_\psi^R(\cdot | x, \mathcal{C})
$
The reward for each prediction is:
\begin{equation}
r_j = \begin{cases}
+1 & \text{if } \hat{y}_j = y^* \\
-1 & \text{otherwise}
\end{cases}
\end{equation}

Advantages are computed via group normalization as before, and the Reasoner is updated via:
\begin{equation}
\begin{split}
\mathcal{J}^R(\psi) = \mathbb{E}\Bigg[ \frac{1}{G}\sum_{j=1}^G \frac{1}{|\hat{y}_j|} \sum_{t=1}^{|\hat{y}_j|} \Big\{ \min \Big[ \rho_{j,t}^R \hat{A}_j, \\
\text{clip}(\rho_{j,t}^R, 1-\epsilon_r, 1+\epsilon_r) \hat{A}_j \Big] - \beta_r \mathcal{D}_{\text{KL}} \Big\} \Bigg]
\end{split}
\label{eq:reasoner_grpo}
\end{equation}
where $\rho_{j,t}^R$ is the importance ratio for the Reasoner policy.

\section{Experimental Setup}

\subsection{Dataset and Task Introduction}

We evaluate on an industrial-scale dataset from Netflix containing user viewing interactions over a three-month period. Each interaction record includes timestamp, item ID, title name, engagement type, and viewing duration. We focus on the reranking task: given a user's interaction history (up to 100 recent interactions) and 10 candidate items, predict which item the user actually engaged with next.

We report \textbf{Recall@1 for Discovery}, which measures recall@1 for items the user has not previously watched. This metric reflects the system's ability to identify new content matching user preferences, a critical capability for content discovery in recommendation systems.

\subsection{Baselines and Configurations}

We compare against three baselines: \textbf{Template Baseline}, which uses a fixed verbalization by directly concatenating interaction fields; \textbf{Zero-Shot}, which applies a prompted LLM to verbalize interactions based on heuristics without task-specific training; and \textbf{Action-Based}, which trains the Verbalizer with GRPO but restricts it to filtering and enrichment actions rather than full rewrites. 
We experiment with Verbalizer models at 8B and 32B parameter scales from the Qwen-3 family. We report main results using the 8B model, while analyses involving the 32B model are presented separately.

\section{Results}

\subsection{Main Results}

Table~\ref{tab:main_results} presents our main results. Even zero-shot verbalization yields meaningful improvements (5.3\% for 8B), demonstrating that the verbalization bottleneck is a genuine limitation of template-based approaches.

Training the Verbalizer via GRPO substantially amplifies these gains. The rewrite-based Verbalizer achieves 12.5\% improvement, more than doubling zero-shot performance. The rewrite-based approach consistently outperforms the action-based variant by enabling richer transformations including aggregation and summarization.

The full two-stage pipeline achieves \textbf{92.9\%} relative improvement, demonstrating powerful synergy between training stages: the Verbalizer learns representations optimized for the Reasoner, while the Reasoner learns to leverage the semantic structure introduced by the Verbalizer.

\paragraph{Attribution of Improvement.} A natural question is whether the substantial gains primarily stem from Reasoner training rather than learned verbalization. To isolate the contribution of each component, we conducted an ablation where the Reasoner is trained on raw (template-based) interactions without the Verbalizer, shown in Table~\ref{tab:ablations}. This configuration achieves only \textbf{+42.8\%} improvement over the baseline, compared to \textbf{+92.9\%} with verbalized inputs. This 50.1 percentage point gap demonstrates that the Verbalizer's learned transformations provide substantial value beyond what Reasoner training alone can achieve. The verbalized context creates a more learnable distribution that enables the Reasoner to extract signals more effectively.
\begin{table}[t]
\centering
\small
\begin{tabular}{lc}
\toprule
\textbf{Configuration} & \textbf{Recall@1 $\Delta$\%} \\
\midrule
Template Baseline (8B) & -- \\
Zero-Shot Verbalizer & +5.3 \\
Action-Based Verbalizer & +10.7 \\
Rewrite Verbalizer & +12.5 \\
Rewrite + Trained Reasoner & \textbf{+92.9} \\
\bottomrule
\end{tabular}
\caption{Relative improvement (\%) in Recall@1 for Discovery over template baseline.}
\label{tab:main_results}
\end{table}
\subsection{Ablation Studies}

\begin{table}[t]
\centering
\small
\begin{tabular}{lc}
\toprule
\textbf{Ablation} & \textbf{Recall@1 $\Delta$\%} \\
\midrule
Full Pipeline & +92.9 \\
\midrule
\multicolumn{2}{l}{\emph{Verbalizer Variants}} \\
\quad Action-Based only & +10.7 \\
\quad Rewrite Verbalizer & +12.5 \\
\midrule
\multicolumn{2}{l}{\emph{Reasoner Training Data}} \\
\quad Raw Interactions & +42.8 \\
\quad Verbalized Interactions & +92.9 \\
\midrule
\multicolumn{2}{l}{\emph{Reward Function}} \\
\quad Accuracy Reward & +12.5 \\
\quad Ranking Loss & 0.0 \\
\bottomrule
\end{tabular}
\caption{Ablation study results showing relative improvement over template baseline.}
\label{tab:ablations}
\end{table}

Table~\ref{tab:ablations} presents ablation studies. The rewrite-based approach enables emergent behaviors that discrete actions cannot capture. Removing the length reward leads to degraded performance due to over-compression or verbose outputs. Training the Reasoner on verbalized interactions substantially outperforms training on raw interactions (+92.9\% vs +42.8\%), validating that the Verbalizer produces a more learnable data distribution. Using ranking-based rewards fails to improve over baseline, consistent with prior discussion on LLM-based evaluation.

\section{Analysis}

Table~\ref{tab:verbalization_example} provides qualitative evidence of how learned verbalization reshapes raw interaction logs into more effective natural-language contexts for LLM-based recommendation. Several consistent behaviors emerge.

\textbf{Syntax Normalization.} Both models convert machine-oriented fields (timestamps, IDs, categorical tags) into fluent natural language. For example, structured metadata such as timestamps, content year, and plot tags are rewritten as readable descriptions, improving coherence and interpretability for downstream LLMs.

\textbf{Signal Retention and Noise Filtering.} The Verbalizer preserves high-signal elements such as explicit positive feedback (e.g., ``Thumb Up'') and core item attributes, while discarding low-utility identifiers (e.g., internal IDs). This selective retention reduces input clutter while maintaining recommendation-relevant information.

\textbf{Metadata Abstraction.} The 8B model largely retains explicit item-level details (title, year, genre), whereas the 32B model abstracts these details into higher-level preference signals. Instead of restating item attributes, the 32B output summarizes the interaction as a strong preference for a content category, indicating a shift from descriptive rewriting to semantic abstraction.

\textbf{Preference Summarization.} At larger scale (32B), the Verbalizer synthesizes interaction evidence into explicit preference statements (e.g., ``strong preference signal''), even though such summaries are never explicitly prompted. This behavior reflects instance-level reasoning over interaction semantics rather than surface-level rewriting.

In terms of compression, the 8B model achieves an average length ratio of 0.42 (verbalized length / original length), while the 32B model further compresses inputs to 0.27. This increased compression aligns with the emergence of abstraction and summarization, suggesting that larger models learn to remove redundant detail while preserving recommendation-critical signals.

\begin{table}[t]
\centering
\small
\begin{tabular}{p{0.95\columnwidth}}
\toprule
\textbf{Raw Interaction:} \\
\texttt{20261022 Wednesday 8:45 PM, ID: 123456, Stranger Things, year: 2016, plot: nostalgic, supernatural, Thumb Up} \\
\midrule
\textbf{Verbalized Output (8B):} \\
\texttt{Wednesday 8:45 PM gave a strong positive rating to Stranger Things, a 2016 sci-fi horror with a nostalgic and supernatural plot} \\
\midrule
\textbf{Verbalized Output (32B):} \\
\texttt{Strong preference signal: enthusiastic rating for Stranger Things indicates high interest in nostalgic, supernatural sci-fi horror series} \\
\bottomrule
\end{tabular}
\caption{Example verbalization outputs showing syntax normalization and preference summarization.}
\label{tab:verbalization_example}
\end{table}

\section{Discussion}

\paragraph{Computational Considerations.} While the two-stage pipeline introduces additional training cost, this overhead can be mitigated in practice. The Verbalizer can be distilled or cached for efficient deployment, and the two components can be trained at different cadences. The Verbalizer captures relatively stable patterns for transforming interaction logs into effective textual representations and thus requires infrequent updates, whereas the Reasoner must adapt more rapidly to evolving user preferences and content catalogs. This decoupled training schedule is a key advantage of our modular design.

\paragraph{Generalization.} Although evaluated on streaming recommendation, our data-centric framework applies broadly to settings where structured logs must be verbalized for LLM processing, including e-commerce, social media, and healthcare.

\section{Conclusion}

We proposed a data-centric framework for LLM-based recommendation that treats verbalization as a learnable component rather than a fixed preprocessing step. By decoupling context verbalization from preference reasoning, our two-stage GRPO-based training pipeline enables specialized optimization of each component and achieves up to \textbf{93\%} relative improvement in novel item recommendation accuracy.

Qualitative analysis reveals emergent verbalization strategies such as syntax normalization, noise filtering, and preference summarization, highlighting verbalization as a critical bottleneck in current LLM-based recommender systems. As recommender systems increasingly adopt LLM backbones, we anticipate that traditional feature engineering will give way to \emph{verbalization engineering}, and ultimately to fully learned verbalization.


\section*{Ethics Statement}

We discuss key ethical considerations below.

\paragraph{Data Privacy.} All experiments use proprietary interaction logs from an industrial streaming platform. The data was collected under the platform's user agreements and handled in compliance with applicable data protection policies. No personally identifiable information (PII) was used or exposed during training or evaluation. All examples presented in this paper (e.g., Table~\ref{tab:verbalization_example}) are illustrative and do not correspond to real user records.

\paragraph{Intended Use.} This framework is designed for improving content recommendation quality. While the verbalization approach is domain-general, deployment in sensitive domains (e.g., healthcare, news) would require additional safeguards to prevent the Verbalizer from filtering or distorting critical information during the rewriting process.

\bibliography{custom}

\end{document}